%% file: main.tex
\documentclass[10pt,twocolumn,letterpaper]{article}

\usepackage[pagenumbers]{cvpr}

\input{preamble}

\usepackage{times}
\usepackage{epsfig}
\usepackage{graphicx}
\usepackage{amsmath}
\usepackage{amssymb}
\usepackage{xcolor}
\usepackage{multirow}
\usepackage{adjustbox}
\usepackage{float}
\usepackage{pifont}
\usepackage{algorithm}
\usepackage{algpseudocode}
\usepackage{tikz}
\usepackage{booktabs}
\usepackage[inkscapelatex=false]{svg}
\usepackage[accsupp]{axessibility}  % Improves PDF readability for those with disabilities.
\usepackage{booktabs}

\definecolor{blue}{RGB}{34,34,225}
\definecolor{myGreen}{RGB}{34, 139, 34}
\definecolor{red}{RGB}{225, 34, 34}
\definecolor{cvprblue}{rgb}{0.21,0.49,0.74}
\usepackage[pagebackref,breaklinks,colorlinks,allcolors=cvprblue]{hyperref}

\DeclareRobustCommand{\proposed}{%
  \texorpdfstring{%
    \BeginAccSupp{method=pdfstringdef,unicode,ActualText={ CoGS }}%
    \texttt{CoGS}%
    \EndAccSupp{}%
  }{CoGS}\xspace%
}

\title{\proposed: Compositional Dynamic Human-Object Scenes \\ Gaussian Splatting from Monocular Video}

\author{Jerrin Bright \quad John Zelek\\
Vision and Image Processing Lab, University of Waterloo, Canada\\
{\tt\small \{jerrin.bright,jzelek\}@uwaterloo.ca}}

\begin{document}
\maketitle

\input{sections/abstract}
\input{sections/intro}
\input{sections/related_work}
\input{sections/method}
\input{sections/experiments}
\input{sections/conclusion}

\newpage

{\small
\bibliographystyle{ieeenat_fullname}
\bibliography{main}
}

% \maketitlesupplementary
% \input{sections/supplementary}

\end{document}

%% file: preamble.tex
\usepackage[table,dvipsnames]{xcolor}
\definecolor{myred}{rgb}{1, 0.7, 0.7}
\definecolor{myorange}{rgb}{1, 0.85, 0.7}
\definecolor{tabfirst}{rgb}{1, 0.7, 0.7}
\definecolor{tabsecond}{rgb}{1, 0.85, 0.7}
\definecolor{tabthird}{rgb}{1, 1, 0.7}
\newcolumntype{C}[1]{>{\centering\arraybackslash}m{#1}}
\newcommand{\tablefirst}[1]{\cellcolor{myred}\textbf{#1}}
\newcommand{\tablesecond}[1]{\cellcolor{myorange}\textbf{#1}}
\newcommand{\gain}[1]{\textcolor[rgb]{0,0,1}{\scriptsize $\uparrow$\,#1}}
\newcommand{\dropmetric}[1]{\textcolor[rgb]{1,0,0}{\scriptsize $\downarrow$\,#1}}

\AtBeginDocument{\DeclareGraphicsRule{.svg}{pdf}{.svg.pdf}{}}

%% file: sections/abstract.tex
\begin{abstract}
Reconstructing dynamic human--object interaction scenes from monocular video is difficult because the human, manipulated object, and background obey different motion models while sharing the same pixels. Existing dynamic radiance-field and Gaussian-splatting methods often entangle these components, causing object motion to leak into the human or static scene, and monocular human reconstruction remains underconstrained in regions that are rarely observed. We present \proposed, a compositional Gaussian-splatting framework for monocular human--object scene reconstruction. \proposed decomposes the video into three coordinated branches: an articulated human initialized from a complete canonical prior, a rigid object field driven by an estimated object trajectory, and a static scene field regularized by weak scene-only planar primitives when available. A six-stage optimization schedule first stabilizes the human and object independently, then fuses them with the scene under full-image supervision, visibility-aware human anchoring, object silhouette and motion constraints, and delayed scene regularization. This design keeps each component responsible for its own geometry and motion while allowing photometric evidence to correct the final composite. Experiments on HOSNeRF and NeuMan show that \proposed improves both human--object interaction reconstruction and in-the-wild human--scene rendering, achieving stronger fidelity and perceptual quality across full-frame and human-focused evaluations. Code will be released upon publication.
\end{abstract}

%% file: sections/intro.tex
\section{Introduction}

Reconstructing a dynamic human, the object they manipulate, and the surrounding scene from monocular video is a tightly coupled inverse problem studied by human-scene and human-object neural rendering methods~\cite{jiang2022neuman,guo2023vid2avatar,liu2023hosnerf}. A single camera provides incomplete geometry, frequent occlusion, and ambiguous depth, while the foreground is not governed by one motion model~\cite{jiang2022neuman,xiang2023occnerf,xiang2024wild2avatar}. The human articulates according to body pose~\cite{loper2023smpl,pavlakos2019expressive}, the manipulated object follows its own motion~\cite{liu2023hosnerf}, and the remaining environment is typically modeled as a static world component observed through the moving camera~\cite{jiang2022neuman,guo2023vid2avatar}. Treating all pixels as one deforming field can fit training views~\cite{park2021nerfies,park2021hypernerf,wu20244d,yang2024deformable}, but in human--object videos this can entangle motions across components, motivating explicit separation of the body, manipulated object, and scene~\cite{liu2023hosnerf}.

Gaussian splatting makes this problem both promising and fragile. Its explicit anisotropic representation supports efficient rendering and direct optimization~\cite{kerbl20233d}, and dynamic Gaussian methods extend it to time-varying monocular scenes~\cite{wu20244d,yang2024deformable,bae2024per,lee2024fully,liu2025modgs}. However, monocular supervision remains underconstrained: human Gaussian/avatar methods rely on articulated priors to stabilize unseen or weakly observed body regions~\cite{kocabas2024hugs,moon2024expressive,xiang2024wild2avatar}, object motion in interaction videos requires object-specific supervision~\cite{liu2023hosnerf}, and depth or geometric priors are often used to stabilize ambiguous scene structure~\cite{liu2025modgs,schonberger2016structure}. A robust monocular method therefore needs more than a single photometric objective.

Our key idea is to reconstruct the scene compositionally. We separate the representation into an articulated human branch, a rigid object branch, and a static scene branch, then train each component only when its supervision is reliable. This separation lets the human prior complete weakly observed body regions, keeps manipulated objects independent from the body and scene, and stabilizes large static surfaces without forcing all pixels into one motion model.

We introduce \proposed, a monocular human--object Gaussian-splatting framework built around this decomposition. Preprocessing aligns COLMAP cameras~\cite{schonberger2016structure}, SMPL motion from monocular pose estimation~\cite{loper2023smpl,patel2024camerahmr}, SAM-based masks~\cite{ravi2024sam2}, object trajectory initialization, canonical human initialization, and optional scene planar primitives into a shared coordinate frame. Optimization then proceeds in six stages: human base fitting, human alpha/boundary refinement, object pose and appearance fitting, object silhouette/rotation refinement, full compositional warmup, and true-resume refinement. This schedule reduces competition between branches and delays aggressive density changes until the corresponding component has stable support.

The resulting representation supports both full composite rendering and branch-level inspection. At inference time, a target camera, human pose, and object pose are rendered by composing the static scene, posed human, and posed object. This enables evaluation on human--object interaction videos from HOSNeRF~\cite{liu2023hosnerf} as well as human--scene benchmarks such as NeuMan~\cite{jiang2022neuman}, where the object branch is not the primary challenge. Across both benchmarks, \proposed improves reconstruction fidelity and perceptual quality, especially in settings with occlusion, moving foreground content, and weakly observed human regions.

The main contributions of this work are as follows:
\begin{itemize}[leftmargin=1.5em]
    \item We introduce a compositional Gaussian-splatting formulation for monocular human--object scene reconstruction, separating the video into an articulated human branch, a rigid manipulated-object branch, and a static scene branch with distinct motion models.
    \item We propose reliability-aware supervision for noisy monocular inputs: visibility-aware anchoring preserves weakly observed human regions.
    \item We design a six-stage optimization curriculum that stabilizes the human and object branches before full compositional fusion with the scene, reducing competition between components that share pixels.
    \item We validate the method on HOSNeRF and NeuMan, showing improved full-frame, human-focused, and human--object reconstruction quality.
\end{itemize}

%% file: sections/related_work.tex
\section{Related Work}

\subsection{Dynamic Gaussian Splatting}

Neural radiance fields established high-quality view synthesis from posed images~\cite{mildenhall2021nerf}, while 3D Gaussian Splatting (3DGS) replaced dense implicit evaluation with an explicit anisotropic Gaussian representation and visibility-aware rasterization for real-time rendering~\cite{kerbl20233d}. Dynamic scene methods extend this idea by modeling time-varying radiance fields with deformation fields, factorized space-time grids, or explicit dynamic Gaussians~\cite{park2021nerfies,park2021hypernerf,wu2022d,fridovich2023k,cao2023hexplane,wu20244d,yang2024deformable,bae2024per,lee2024fully,liu2025modgs}. These methods are effective for generic deforming scenes, but monocular human--object videos contain components with different physical constraints: articulated body motion, approximately rigid object motion, and mostly static scene structure. \proposed therefore uses separate human, object, and scene branches rather than asking a single dynamic field to explain all pixels.

\subsection{Human Gaussian Splatting}

Human-specific neural rendering uses body models such as SMPL/SMPL-X to factor pose from appearance~\cite{loper2023smpl,pavlakos2019expressive, bright2024distribution}. NeRF-based methods such as Neural Body, HumanNeRF, NeuMan, and Vid2Avatar combine articulated human priors with radiance fields for monocular free-viewpoint rendering~\cite{peng2021neural,weng2022humannerf,jiang2022neuman,guo2023vid2avatar}. More recent Gaussian approaches improve efficiency and editability by learning canonical human Gaussians that are deformed into posed frames~\cite{kocabas2024hugs,hu2024gauhuman,hu2024gaussianavatar,qian20243dgs, bright2025gen4d}. Human-scene methods further reconstruct the moving person and static environment jointly~\cite{xue2024hsr,zhang2025odhsr}. These methods retain animatability and achieve strong rendering quality, but monocular training still leaves persistently hidden body regions underconstrained, especially when objects or scene structure occlude the body. Occlusion-aware methods address this with explicit occluder or visibility modeling~\cite{xiang2023occnerf,xiang2024wild2avatar}; \proposed instead uses a complete canonical human prior and modulates its influence by per-Gaussian visibility, allowing observed regions to adapt while preserving plausible completion in unobserved areas.

% Complementary work improves the human evidence used by such pipelines. Distribution and depth-aware transformers use scene-depth and distribution cues for robust monocular human mesh recovery~\cite{bright2024distribution}, while Gen4D and Avatar4D synthesize diverse 4D humans, scenes, and domain-specific motions for human-centric pose estimation and transfer~\cite{bright2025gen4d,bright2025avatar4d}. These methods strengthen pose and data priors; \proposed uses such human initialization cues only as support for sequence-specific compositional reconstruction.
Overall, human Gaussian methods show that body-model guidance and canonical human Gaussians are strong priors for animatable reconstruction, but monocular videos still require careful handling of occlusion and weakly observed regions. This motivates the visibility-aware human branch in \proposed: the canonical prior supports initialization and hidden-region completion, while sequence-specific photometric and mask evidence remains responsible for correcting the visible body.

\subsection{Human--Object Reconstruction}

Human--object reconstruction must recover body motion, object geometry, contact, and occlusion ordering. Datasets such as BEHAVE and ARCTIC provide controlled supervision for full-body or hand-object interactions~\cite{bhatnagar2022behave,fan2023arctic}. HOSNeRF reconstructs dynamic human--object--scene radiance fields from a monocular video~\cite{liu2023hosnerf}, while recent approaches reconstruct category-agnostic hand--object or human--object interactions from video~\cite{fan2024hold,mir2025gaspacho}. \proposed keeps the manipulated object as an independent Gaussian branch with its own rigid trajectory, silhouette supervision, and motion regularization, so moving foreground pixels can be explained without corrupting the articulated human or static scene representation.

\subsection{Prior-Guided and Scene-Regularized Optimization}

Priors are widely used in inverse graphics to stabilize reconstruction in weakly observed regions. Parametric body models constrain articulation~\cite{loper2023smpl,pavlakos2019expressive}, learned clothed-human priors improve single-view completion~\cite{saito2019pifu,saito2020pifuhd,xiu2022icon,xiu2023econ}, and generative or geometric priors can initialize object and scene structure~\cite{tang2023dreamgaussian,jiang2022neuman,guo2023vid2avatar}. \proposed combines these ideas in a targeted way: a visibility-weighted human prior anchors canonical body Gaussians only where video evidence is weak, while planar primitives weakly regularize large static surfaces after warmup. These priors are not treated as fixed outputs; they provide structure that can be corrected by sequence-level photometric, mask, and motion evidence during staged optimization.

%% file: sections/method.tex
\begin{figure}[t]
\centering
\includegraphics[width=\columnwidth]{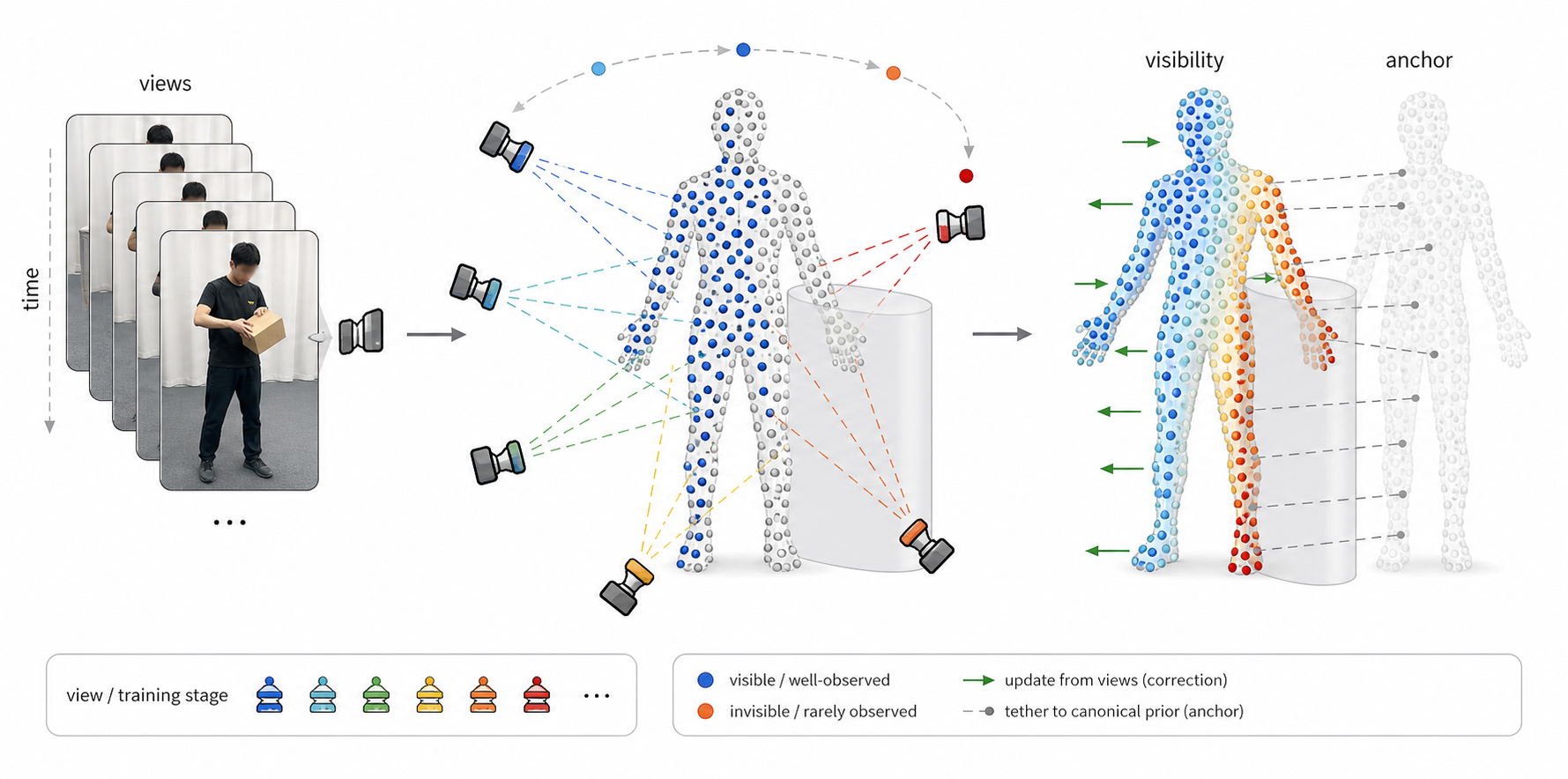}
\caption{\textbf{Visibility-aware human anchoring.}
Well-observed Gaussians (blue) are corrected by image evidence, while rarely observed or occluded regions (orange/red) remain softly tethered to the canonical prior.}
\label{fig:visibility_anchor}
\end{figure}

\begin{figure*}[t]
\centering
\includegraphics[width=\textwidth]{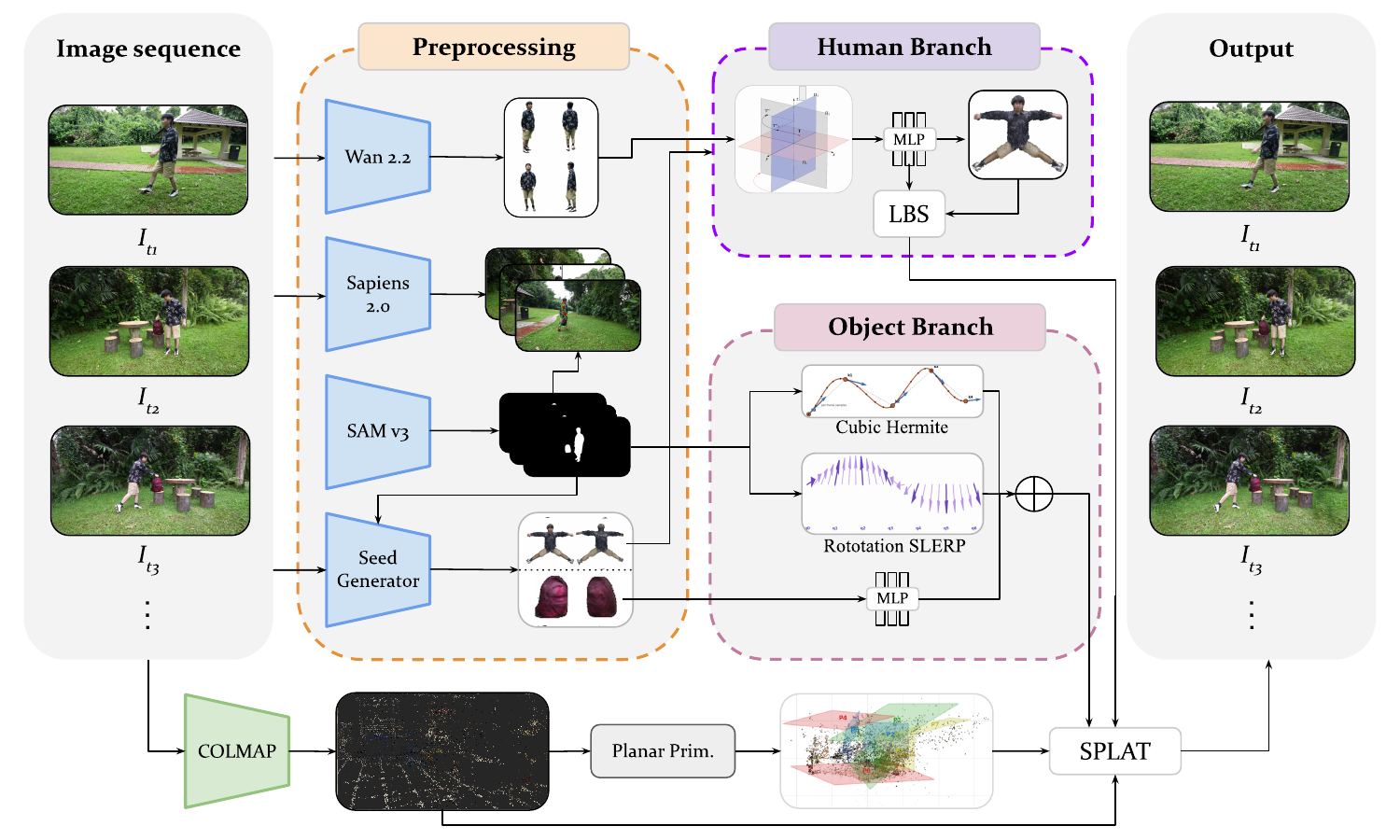}
\caption{\textbf{Overview of \proposed: a trust-then-correct compositional pipeline.} From a monocular human--object video, preprocessing yields masks, COLMAP camera calibration and sparse geometry, SMPL motion, an object trajectory, optional scene planar primitives, and a complete canonical human seed. The \textbf{human branch} decodes a canonical Gaussian human from a triplane--MLP field and poses it by linear blend skinning (LBS) under SMPL body motion. The \textbf{object branch} models the manipulated object as a rigid Gaussian set whose per-frame pose comes from a compact keyframed trajectory---\emph{cubic-Hermite} interpolation for translation and spherical linear interpolation (\emph{SLERP}) for rotation---which is then fused with the other branches. The \textbf{scene branch} is a static Gaussian field initialized from COLMAP with weak planar primitives. A shared splatting renderer composites the posed human, object, and scene to reconstruct each frame. The trust-then-correct schedule preserves the human prior in rarely observed regions through visibility-aware anchoring, while well-observed regions, object motion, and background structure are corrected by image evidence.}
\label{fig:overview}
\end{figure*}

\section{Method}
\label{sec:method}

\subsection{Problem Setup and Notation}
\label{sec:problem_setup}

Given a monocular video, our goal is to reconstruct a compositional Gaussian scene that separates the static environment, the articulated human, and the manipulated object. Fig.~\ref{fig:overview} summarizes the resulting preprocessing, branch-specific modeling, and compositional rendering pipeline. This decomposition is necessary because the three components follow different motion models: the background is fixed in the world frame, the human is driven by SMPL-based body motion, and the object follows its own estimated rigid trajectory. At each training frame, the renderer composites the scene, posed human, and posed object under the calibrated camera, and the optimization supervises the resulting image with RGB, mask, silhouette, motion, and regularization losses.

The optimizer receives the following aligned signals for each frame $t$: RGB image $I_t$, calibrated camera $\mathcal{C}_t$, human mask $M^H_t$, object mask $M^O_t$, SMPL pose/shape $(\theta_t,\beta)$, and an initial rigid object pose $\bar{T}^O_t=(\bar{R}^O_t,\bar{\mathbf{t}}^O_t)$. It also receives a canonical human seed $\mathcal{G}^H_0=\{(\boldsymbol{\mu}^0_i,\mathbf{s}^0_i,\mathbf{q}^0_i,\alpha^0_i,\mathbf{c}^0_i)\}_i$, where each Gaussian stores position, scale, rotation, opacity, and color, plus optional scene planes $\mathcal{P}=\{(\mathbf{n}_j,d_j)\}_j$. Preprocessing produces these quantities from the raw video using off-the-shelf reconstruction, segmentation, pose-estimation, and plane-extraction modules, but the optimizer treats them as initialization or soft supervision rather than immutable ground truth. We keep only these optimizer interfaces in the main paper; the full preprocessing pipeline is detailed in supplementary Sec.~2. All signals are expressed in one world coordinate frame before staged optimization begins, so the human, object, and scene branches can be trained independently and then composed without changing coordinate conventions.

Because these preprocessed signals can be noisy, \proposed treats them as soft initialization and supervision: masks affect losses and visibility estimates, camera/pose estimates define the shared coordinate frame, object seeds initialize geometry and motion, and planar primitives are optional weak priors. Moderate noise is corrected during staged optimization through photometric, silhouette, motion, and visibility-aware losses; severe camera, segmentation, pose, or object-seed failure can still degrade the corresponding branch.

\subsection{Human Gaussian Branch}
\label{sec:human_branch}

The human branch represents the person as a canonical Gaussian set deformed by the SMPL motion estimated during preprocessing. We initialize this canonical set from the human-prior seed instead of relying only on monocular fitting. The seed provides complete geometry, opacity, and coarse appearance for both visible and persistently hidden body regions, while the SMPL-driven deformation pathway keeps the representation animatable. Each Gaussian is attached to the SMPL body using surface-based skinning weights, so frame-level body motion can pose the canonical human without changing the canonical parameterization.

The key design is visibility-aware correction. For canonical human Gaussian $i$, we pose it into frame $t$, project it to pixel $\mathbf{u}_{it}$, and mark it supported when it lands inside the human mask, outside the object mask, and contributes non-negligible human opacity. Its visibility score is
\[
\begin{aligned}
v_i&=\frac{1}{|\mathcal{T}|}\sum_{t\in\mathcal{T}}\chi_{it},\\
\chi_{it}
&=\mathbf{1}\!\left[\mathbf{u}_{it}\in M^H_t\right]
\mathbf{1}\!\left[\mathbf{u}_{it}\notin M^O_t\right]
\mathbf{1}\!\left[\alpha^H_{it}>\epsilon\right],
\end{aligned}
\]
where the product of indicators makes $\chi_{it}=1$ only when all support tests pass. We use the detached weight $w_i=1-v_i$ to anchor weakly observed Gaussians to the canonical seed:
\[
\begin{aligned}
\mathcal{L}_{\mathrm{anchor}}
=\sum_i w_i\big(
&\|\boldsymbol{\mu}_i-\boldsymbol{\mu}^0_i\|_1
+\lambda_s\|\mathbf{s}_i-\mathbf{s}^0_i\|_1\\
&+\lambda_q d_R(\mathbf{q}_i,\mathbf{q}^0_i)
+\lambda_\alpha|\alpha_i-\alpha^0_i|
\big),
\end{aligned}
\]
where $d_R$ is the geodesic distance between rotations. Thus frequently visible Gaussians can move toward photometric and mask evidence, while weakly observed Gaussians remain softly anchored to the seed. Fig.~\ref{fig:visibility_anchor} illustrates this view-dependent anchoring.

Human training uses RGB reconstruction, mask/silhouette supervision, alpha and boundary losses, human-separated rendering terms, Gaussian regularization, and the visibility-aware anchor. We intentionally delay aggressive density changes for the human branch, because early densification can destroy the prior structure before the masks and appearance have stabilized.

\subsection{Object Gaussian Branch}
\label{sec:object_branch}

The object branch models the manipulated object as a separate Gaussian field with its own rigid motion trajectory. This is important because treating the object as either part of the human or part of the static scene causes leakage: the human branch may absorb object pixels near contact, while the scene branch may bake a moving object into the background. We initialize the object from masks $M^O_t$, COLMAP sparse geometry, available depth cues, and the silhouette-fit trajectory $\bar{T}^O_t$ in the same world frame as the cameras and SMPL body. The initial trajectory is used as a soft pose prior,
\[
\mathcal{L}^{O}_{\mathrm{pose}}
=\sum_t \|\mathbf{t}^O_t-\bar{\mathbf{t}}^O_t\|_1
+\lambda_R d_R(\mathbf{q}^O_t,\bar{\mathbf{q}}^O_t),
\]
where $\bar{\mathbf{q}}^O_t$ encodes $\bar{R}^O_t$. Thus object pose errors are refined during training rather than frozen from preprocessing.

\paragraph{Rigid keyframed motion.}
The object is rigid, so rather than an unconstrained per-frame pose we parametrize its motion with a small set of $K$ keyframes and interpolate between them; this acts as a strong temporal regularizer for a small, low-contrast object whose per-frame mask signal is noisy. Each keyframe stores a translation $\mathbf{p}_k$, a velocity $\mathbf{v}_k$, and a unit quaternion $\mathbf{q}_k$. For frame $\tau$ we map time onto a segment $[k_0,k_1]$ with local parameter $u\in[0,1]$ and obtain the translation from a \emph{cubic-Hermite} spline,
\begin{equation}
\mathbf{t}(\tau)=h_{00}\,\mathbf{p}_{k_0}+h_{10}\,\mathbf{v}_{k_0}+h_{01}\,\mathbf{p}_{k_1}+h_{11}\,\mathbf{v}_{k_1},
\end{equation}
with Hermite basis
\[
\begin{gathered}
h_{00}=2u^3-3u^2+1,\quad h_{10}=u^3-2u^2+u,\\
h_{01}=-2u^3+3u^2,\quad h_{11}=u^3-u^2 .
\end{gathered}
\]
Using the learned endpoint velocities yields a $C^1$-continuous, smoothly accelerating path instead of the piecewise-linear, velocity-discontinuous motion of naive interpolation. The rotation is obtained by spherical linear interpolation (\emph{SLERP}) of the keyframe quaternions along the shortest arc, which stays on the unit-quaternion manifold and avoids the gimbal lock and non-uniform angular speed of Euler-angle or matrix interpolation. The resulting rigid pose is applied to every object Gaussian as
\[
\mathbf{x}(\tau)=
\mathbf{R}\big(\mathbf{q}(\tau)\big)
\big(\mathbf{x}_{\mathrm{canon}}-\mathbf{c}\big)
+\mathbf{t}(\tau),
\]
i.e.\ the canonical object is rotated about its center $\mathbf{c}$ and translated, while its canonical appearance (explicit per-Gaussian spherical-harmonic color) is shared across all frames. Training jointly refines the per-Gaussian attributes and the keyframes $\{\mathbf{p}_k,\mathbf{v}_k,\mathbf{q}_k\}$, regularized by acceleration and rotation-smoothness terms and a prior anchoring the trajectory to its silhouette-fit initialization.

Object optimization is also visibility-driven, but the role of visibility differs from the human branch. The object is supervised most reliably when its silhouette is clean and it is not heavily occluded by the body. We therefore emphasize silhouette, edge, pose, and rotation-smoothness terms early, and then refine appearance once the pose and support are stable. Object densification is bounded and delayed so that pose errors are corrected through motion refinement rather than hidden by uncontrolled Gaussian growth.

This separation lets the object preserve a compact geometry and consistent motion through contact-heavy frames. During joint rendering, the object branch can explain moving foreground pixels without forcing the human seed or static scene to compensate for object motion.

\subsection{Scene Gaussian Branch}
\label{sec:plane_method}

The scene branch captures the static background and any non-moving environment structure. It is initialized from the COLMAP camera and sparse-point reconstruction and optimized with standard Gaussian photometric supervision. Unlike the human and object branches, the scene does not require articulated or rigid per-frame deformation, but it must remain stable under occlusion from the dynamic foreground.

To improve scene geometry in low-texture regions, we extract planar primitives from COLMAP-aligned scene/depth geometry when available and use them as weak structural priors. These planes mainly support floors, walls, and large background surfaces where photometric gradients alone are ambiguous. Rather than forcing all nearby Gaussians to lie exactly on a plane, we assign only compatible scene Gaussians to nearby planes and penalize their point-to-plane distance after warmup:
\[
\mathcal{L}_{\mathrm{plane}}
=\sum_{(i,j)\in\mathcal{A}_{\mathrm{plane}}}
\rho(\mathbf{n}_j^\top\boldsymbol{\mu}^{S}_i+d_j),
\]
where $\mathcal{A}_{\mathrm{plane}}$ contains scene-only Gaussian--plane assignments and $\rho$ is a robust penalty. This keeps the prior useful for regularizing large surfaces without overriding image evidence or early density allocation.

The scene branch is therefore trained as a static Gaussian field with delayed geometric regularization. It provides the base layer for the final composite renderer, while the human and object branches explain dynamic foreground content.

\subsection{Compositional Fusion}
\label{sec:compositional_fusion}

After the component branches are stable, we compose the scene, human, and object fields in a shared renderer. The scene acts as the static base, while posed human and object Gaussians are rendered in the camera frame using their masks, visibility estimates, and occlusion ordering. This compositional design avoids a single Gaussian cloud trying to explain mutually incompatible motions.

The fusion stage is deliberately lightweight in formulation: it reuses the same branch-specific representations and supervises the final rendered image with full-frame reconstruction, foreground consistency, object motion regularization, visibility-aware human anchoring, and delayed scene planar regularization. The exact training schedule for moving from isolated branches to the fused renderer is described next in the six-stage optimization strategy.

\begin{table*}[!t]
\centering
\small
\setlength\tabcolsep{1.0pt}
\caption{
\textbf{HOSNeRF quantitative comparison.} Per-scene quantitative evaluation on the \textbf{HOSNeRF}~\cite{liu2023hosnerf} dataset against reproducible dynamic-scene and human-scene baselines.}
\label{tab:hosnerf_comparison}
\resizebox{\textwidth}{!}{%
\begin{tabular}{C{4.3cm}|C{1.1cm}C{1.1cm}|C{1.1cm}C{1.1cm}|C{1.1cm}C{1.1cm}|C{1.1cm}C{1.1cm}|C{1.1cm}C{1.1cm}|C{1.1cm}C{1.1cm}}
\specialrule{.1em}{.05em}{.05em}
\multirow{2}{*}{Methods} &
\multicolumn{2}{c|}{\textbf{Backpack}} &
\multicolumn{2}{c|}{\textbf{Tennis}} &
\multicolumn{2}{c|}{\textbf{Suitcase}} &
\multicolumn{2}{c|}{\textbf{Playground}} &
\multicolumn{2}{c|}{\textbf{Dance}} &
\multicolumn{2}{c}{\textbf{Lounge}} \\
& PSNR$\uparrow$ & LPIPS$\downarrow$
& PSNR$\uparrow$ & LPIPS$\downarrow$
& PSNR$\uparrow$ & LPIPS$\downarrow$
& PSNR$\uparrow$ & LPIPS$\downarrow$
& PSNR$\uparrow$ & LPIPS$\downarrow$
& PSNR$\uparrow$ & LPIPS$\downarrow$ \\ \hline
K-Planes~\cite{fridovich2023k} 
& 19.05 & 0.557 
& 19.31 & 0.536 
& 18.64 & 0.602 
& 17.92 & 0.635 
& 18.17 & 0.623 
& 24.21 & 0.453 \\
D$^2$NeRF~\cite{wu2022d} 
& 20.52 & 0.608 
& 23.97 & 0.540 
& 20.99 & 0.645 
& 21.23 & 0.616 
& 19.92 & 0.647 
& 27.13 & 0.509 \\
Nerfies~\cite{park2021nerfies} 
& 19.56 & 0.559 
& 22.12 & 0.443 
& 19.01 & 0.555 
& 21.14 & 0.533 
& 19.37 & 0.524 
& 25.90 & 0.342 \\
HyperNeRF~\cite{park2021hypernerf} 
& 19.62 & 0.587 
& 21.26 & 0.510 
& 19.41 & 0.607 
& 21.67 & 0.578 
& 19.30 & 0.601 
& 27.25 & 0.332 \\
NeuMan~\cite{jiang2022neuman} 
& 21.21 & 0.478 
& 23.17 & 0.442 
& 20.84 & 0.551 
& 21.46 & 0.551 
& 21.19 & 0.490 
& 28.40 & 0.341 \\
4DGS~\cite{wu20244d} 
& 24.49 & 0.192 
& \tablesecond{26.57} & 0.162 
& 17.98 & 0.460 
& 24.34 & 0.222 
& 21.34 & 0.212 
& \tablesecond{30.50} & 0.067 \\
D3DGS~\cite{yang2024deformable} 
& 24.06 & \tablefirst{0.099}
& 25.09 & \tablesecond{0.125}
& 17.85 & 0.453 
& 23.93 & 0.141 
& 21.07 & \tablesecond{0.117}
& 26.90 & 0.072 \\
E-D3DGS~\cite{bae2024per} 
& \tablesecond{24.78} & 0.146 
& 26.53 & 0.161 
& 18.05 & 0.461 
& 24.37 & 0.206 
& \tablesecond{23.87} & 0.159 
& 30.04 & 0.086 \\
Ex4DGS~\cite{lee2024fully} 
& 18.07 & 0.433 
& 17.90 & 0.399 
& 15.25 & 0.557 
& 16.36 & 0.535 
& 17.08 & 0.529 
& 23.15 & 0.310 \\
ExAvatar~\cite{moon2024expressive} 
& 24.15 & 0.107 
& 23.57 & 0.160 
& 20.32 & \tablesecond{0.260}
& \tablesecond{25.30} & \tablesecond{0.129}
& 23.32 & 0.170 
& 29.43 & \tablesecond{0.048}\\
HOSNeRF~\cite{liu2023hosnerf} 
& 22.56 & 0.243 
& 24.15 & 0.320 
& \tablesecond{21.74} & 0.382 
& 22.67 & 0.336 
& 22.63 & 0.248 
& 27.74 & 0.227 \\
% & \tablefirst{25.78} & \tablefirst{0.082}
% & \tablefirst{27.12} & \tablesecond{0.108}
% & \tablesecond{22.09} & \tablesecond{0.246}
% & 25.23 & \tablefirst{0.103}
% & \tablesecond{24.17} & \tablesecond{0.098}
% & \tablesecond{30.97} & \tablesecond{0.048}\\
\hline
\textbf{Ours (\proposed)}
& \tablefirst{25.34} & \tablesecond{0.10}
& \tablefirst{26.97} & \tablefirst{0.07}
& \tablefirst{24.33} & \tablefirst{0.12}
& \tablefirst{25.53} & \tablefirst{0.12}
& \tablefirst{25.13} & \tablefirst{0.09}
& \tablefirst{31.07} & \tablefirst{0.03}\\
& \gain{+2.3\%} & \dropmetric{-1.0\%}
& \gain{+1.5\%} & \gain{+44.0\%}
& \gain{+11.9\%} & \gain{+53.8\%}
& \gain{+0.9\%} & \gain{+7.0\%}
& \gain{+5.3\%} & \gain{+23.1\%}
& \gain{+1.9\%} & \gain{+37.5\%}\\
\specialrule{.1em}{.05em}{.05em}
\end{tabular}%
}
\vspace{-2mm}
\end{table*}

\section{Optimization}
\label{sec:optimization}
\label{sec:training_curriculum}

A staged schedule is essential because the human, object, and scene branches compete for the same pixels. We therefore keep the full curriculum but present it as three stabilization phases; the exact objective terms and density-control settings are listed in supplementary Table~4.

\paragraph{Human stabilization (Stages I--II).}
We train the human branch alone from $0$--$4$k iterations using the prior seed, RGB/mask supervision, human-separated rendering, unpremultiplied appearance, and visibility anchoring, while disabling human densification and pruning. We then resume to $5$k iterations with stronger alpha, BCE/Dice, and edge terms to sharpen the body support before any object or scene component can compete for the same foreground pixels.

\paragraph{Object stabilization (Stages III--IV).}
With the polished human frozen, the object branch is optimized from $0$--$4$k iterations from the preprocessed trajectory, masks, COLMAP sparse geometry, and dense-depth cues when available. Silhouette-dominant supervision first corrects pose geometrically; a short $4$--$6$k polish phase adds edge, unpremultiplied RGB, rotation-smoothness, and rotation-prior terms while keeping object densification delayed and bounded.

\paragraph{Compositional training (Stages V--VI).}
We load the stabilized human and object, initialize a fresh scene field, and train the full renderer from $0$--$5.5$k iterations with scene densification enabled and human/object topology changes kept conservative. A final true-resume pass to $15$k iterations reduces aggressive resets and refines full-image appearance, alpha consistency, object motion, visibility anchoring, and delayed scene-planar regularization.

%% file: sections/experiments.tex
\section{Experiments}

\begin{table*}[t]
    \centering
\small
\setlength\tabcolsep{1.0pt}
    \caption{\textbf{NeuMan full-frame comparison.} Comparison with previous work on test images of the NeuMan dataset~\cite{jiang2022neuman} using PSNR, SSIM and LPIPS metrics. \proposed achieves the best performance across all reported scenes and metrics. We color code each cell as \colorbox{myred}{\textbf{best}} and \colorbox{myorange}{\textbf{second best}.}}
    \label{tab:neuman_human_scene}
    \resizebox{\textwidth}{!}{%
    \begin{tabular}{c|ccc|ccc|ccc|ccc|ccc|ccc}
    \toprule
        & \multicolumn{3}{c|}{\textbf{Seattle}} & \multicolumn{3}{c|}{\textbf{Citron}} & \multicolumn{3}{c|}{\textbf{Parking}} & \multicolumn{3}{c|}{\textbf{Bike}} & \multicolumn{3}{c|}{\textbf{Jogging}} & \multicolumn{3}{c}{\textbf{Lab}}   \\
    \midrule
        & PSNR $\uparrow$ & SSIM $\uparrow$ & LPIPS $\downarrow$ & PSNR $\uparrow$ & SSIM $\uparrow$ & LPIPS $\downarrow$ & PSNR $\uparrow$ & SSIM $\uparrow$ & LPIPS $\downarrow$ & PSNR $\uparrow$ & SSIM $\uparrow$ & LPIPS $\downarrow$ & PSNR $\uparrow$ & SSIM $\uparrow$ & LPIPS $\downarrow$ & PSNR $\uparrow$ & SSIM $\uparrow$ & LPIPS $\downarrow$  \\
    \midrule
    NeRF-T & 21.84 & 0.69 & 0.37 & 12.33 & 0.49 & 0.65 & 21.98 & 0.69 & 0.46 & 21.16 & 0.71 & 0.36 & 20.63 & 0.53 & 0.49 & 20.52 & 0.75 & 0.39 \\
    HyperNeRF & 16.43 & 0.43 & 0.40 & 16.81 & 0.41 & 0.56 & 16.04 & 0.38 & 0.62 & 17.64 & 0.42 & 0.43 & 18.52 & 0.39 & 0.52 & 16.75 & 0.51 & 0.23 \\
    Vid2Avatar & 17.41 & 0.56 & 0.60 & 14.32 & 0.62 & 0.65 & 21.56 & 0.69 & 0.50 & 14.86 & 0.51 & 0.69 & 15.04 & 0.41 & 0.70 & 13.96 & 0.60 & 0.68 \\
    NeuMan & \cellcolor{tabthird}23.99 & \cellcolor{tabthird}0.78 & \cellcolor{tabthird}0.26 & \cellcolor{tabthird}24.63 & \cellcolor{tabthird}0.81 & \cellcolor{tabthird}0.26 & \cellcolor{tabthird}25.43 & \cellcolor{tabthird}0.80 & \cellcolor{tabthird}0.31 & \cellcolor{tabsecond}25.55 & \cellcolor{tabthird}0.83 & \cellcolor{tabthird}0.23 & \cellcolor{tabthird}22.70 & \cellcolor{tabthird}0.68 & \cellcolor{tabthird}0.32 & \cellcolor{tabthird}24.96 & \cellcolor{tabthird}0.86 & \cellcolor{tabthird}0.21 \\
    HUGS & \cellcolor{tabsecond}25.94 & \cellcolor{tabsecond}0.85 & \cellcolor{tabsecond}0.13 & \cellcolor{tabsecond}25.54 & \cellcolor{tabsecond}0.86 & \cellcolor{tabsecond}0.15 & \cellcolor{tabsecond}26.86 & \cellcolor{tabsecond}0.85 & \cellcolor{tabsecond}0.22 & \cellcolor{tabthird}25.46 & \cellcolor{tabsecond}0.84 & \cellcolor{tabsecond}0.13 & \cellcolor{tabsecond}23.75 & \cellcolor{tabsecond}0.78 & \cellcolor{tabsecond}0.22 & \cellcolor{tabsecond}26.00 & \cellcolor{tabsecond}0.92 & \cellcolor{tabsecond}0.09 \\
    \midrule
    Ours (\proposed) & \cellcolor{tabfirst}29.18 & \cellcolor{tabfirst}0.89 & \cellcolor{tabfirst}0.07 & \cellcolor{tabfirst}28.37 & \cellcolor{tabfirst}0.90 & \cellcolor{tabfirst}0.06 & \cellcolor{tabfirst}31.28 & \cellcolor{tabfirst}0.87 & \cellcolor{tabfirst}0.10 & \cellcolor{tabfirst}29.40 & \cellcolor{tabfirst}0.91 & \cellcolor{tabfirst}0.05 & \cellcolor{tabfirst}25.76 & \cellcolor{tabfirst}0.82 & \cellcolor{tabfirst}0.13 & \cellcolor{tabfirst}29.31 & \cellcolor{tabfirst}0.93 & \cellcolor{tabfirst}0.05 \\
    & \gain{+12.49\%} & \gain{+4.71\%} & \gain{+46.15\%} & \gain{+11.08\%} & \gain{+4.65\%} & \gain{+60.00\%} & \gain{+16.46\%} & \gain{+2.35\%} & \gain{+54.55\%} & \gain{+15.07\%} & \gain{+8.33\%} & \gain{+61.54\%} & \gain{+8.46\%} & \gain{+5.13\%} & \gain{+40.91\%} & \gain{+12.73\%} & \gain{+1.09\%} & \gain{+44.44\%} \\
    \bottomrule
\end{tabular}%
    }
\end{table*}

\begin{table*}[t]
\centering
\small
\setlength\tabcolsep{1.2pt}
\caption{\textbf{NeuMan human-only comparison.} Comparison with previous work on the NeuMan dataset~\cite{jiang2022neuman} over \textbf{human-only} regions cropped using a tight bounding box. Performance is evaluated on PSNR, SSIM and LPIPS metrics.}
\label{tab:neuman_human}
\resizebox{\textwidth}{!}{%
\begin{tabular}{c|ccc|ccc|ccc|ccc|ccc|ccc}
\toprule
    & \multicolumn{3}{c|}{\textbf{Seattle}} & \multicolumn{3}{c|}{\textbf{Citron}} & \multicolumn{3}{c|}{\textbf{Parking}} & \multicolumn{3}{c|}{\textbf{Bike}} & \multicolumn{3}{c|}{\textbf{Jogging}} & \multicolumn{3}{c}{\textbf{Lab}}   \\
\midrule
    & PSNR $\uparrow$ & SSIM $\uparrow$ & LPIPS $\downarrow$ & PSNR $\uparrow$ & SSIM $\uparrow$ & LPIPS $\downarrow$ & PSNR $\uparrow$ & SSIM $\uparrow$ & LPIPS $\downarrow$ & PSNR $\uparrow$ & SSIM $\uparrow$ & LPIPS $\downarrow$ & PSNR $\uparrow$ & SSIM $\uparrow$ & LPIPS $\downarrow$ & PSNR $\uparrow$ & SSIM $\uparrow$ & LPIPS $\downarrow$ \\
\midrule
Vid2Avatar & 16.90 & 0.51 & 0.27 & 15.96 & 0.59 & 0.28 & \cellcolor{tabthird}18.51 & 0.65 & 0.26 & 12.44 & 0.39 & 0.54 & 16.36 & 0.46 & 0.30 & 15.99 & 0.62 & 0.34 \\
NeuMan & \cellcolor{tabthird}18.42 & \cellcolor{tabthird}0.58 & \cellcolor{tabthird}0.20 & \cellcolor{tabthird}18.39 & \cellcolor{tabthird}0.64 & \cellcolor{tabthird}0.19 & 17.66 & \cellcolor{tabthird}0.66 & \cellcolor{tabthird}0.24 & \cellcolor{tabthird}19.05 & \cellcolor{tabthird}0.66 & \cellcolor{tabthird}0.21 & \cellcolor{tabsecond}17.57 & \cellcolor{tabthird}0.54 & \cellcolor{tabthird}0.29 & \cellcolor{tabthird}18.76 & \cellcolor{tabthird}0.73 & \cellcolor{tabthird}0.23 \\
HUGS & \cellcolor{tabsecond}19.06 & \cellcolor{tabsecond}0.67 & \cellcolor{tabsecond}0.15 & \cellcolor{tabsecond}19.16 & \cellcolor{tabsecond}0.71 & \cellcolor{tabsecond}0.16 & \cellcolor{tabsecond}19.44 & \cellcolor{tabsecond}0.73 & \cellcolor{tabsecond}0.17 & \cellcolor{tabsecond}19.48 & \cellcolor{tabsecond}0.67 & \cellcolor{tabsecond}0.18 & \cellcolor{tabthird}17.45 & \cellcolor{tabsecond}0.59 & \cellcolor{tabsecond}0.27 & \cellcolor{tabsecond}18.79 & \cellcolor{tabsecond}0.76 & \cellcolor{tabsecond}0.18 \\
\midrule
Ours (\proposed) & \cellcolor{tabfirst}24.95 & \cellcolor{tabfirst}0.92 & \cellcolor{tabfirst}0.04 & \cellcolor{tabfirst}26.51 & \cellcolor{tabfirst}0.92 & \cellcolor{tabfirst}0.03 & \cellcolor{tabfirst}27.67 & \cellcolor{tabfirst}0.91 & \cellcolor{tabfirst}0.04 & \cellcolor{tabfirst}25.19 & \cellcolor{tabfirst}0.86 & \cellcolor{tabfirst}0.06 & \cellcolor{tabfirst}23.01 & \cellcolor{tabfirst}0.82 & \cellcolor{tabfirst}0.09 & \cellcolor{tabfirst}26.78 & \cellcolor{tabfirst}0.92 & \cellcolor{tabfirst}0.05 \\
& \gain{+30.90\%} & \gain{+37.31\%} & \gain{+73.33\%} & \gain{+38.36\%} & \gain{+29.58\%} & \gain{+81.25\%} & \gain{+42.34\%} & \gain{+24.66\%} & \gain{+76.47\%} & \gain{+29.31\%} & \gain{+28.36\%} & \gain{+66.67\%} & \gain{+30.96\%} & \gain{+38.98\%} & \gain{+66.67\%} & \gain{+42.52\%} & \gain{+21.05\%} & \gain{+72.22\%} \\
\bottomrule
\end{tabular}%
}
\vspace{-2mm}
\end{table*}

\subsection{Datasets}
\label{sec:datasets}

We use HOSNeRF and NeuMan to evaluate complementary aspects of the problem: human--object interaction and in-the-wild human--scene reconstruction. For HOSNeRF, following HOSNeRF~\cite{liu2023hosnerf}, we uniformly select 16 frames per sequence for testing and utilize the remainder for training. For NeuMan, we follow the HUGS evaluation protocol and report results on the held-out test images provided by the benchmark.

\noindent\textbf{HOSNeRF.} A monocular dynamic-scene dataset capturing human--object interactions across 6 indoor/outdoor locations with 6 subjects. Each sequence contains 300--400 frames.

\noindent\textbf{NeuMan~\cite{jiang2022neuman}.} A monocular human--scene benchmark built from in-the-wild videos with a moving camera and a dynamic person embedded in a static scene. Following HUGS, we evaluate on six public sequences: \textit{Seattle}, \textit{Citron}, \textit{Parking}, \textit{Bike}, \textit{Jogging}, and \textit{Lab}. NeuMan complements HOSNeRF because it stresses full human--scene reconstruction under real camera motion, scale variation, and partial body visibility, making it a direct test of whether the human prior improves both complete-scene rendering and human-only regions.

\subsection{Training Details}
\label{sec:training_details}

All experiments follow the six-stage schedule in Sec.~\ref{sec:training_curriculum}. After independent human and object fitting, we load both branches into a full compositor with a fresh scene field; density control is conservative for the human prior, delayed and bounded for the object, and enabled for the scene during compositional training. At inference, held-out frames are rendered by composing the static scene, posed human, and posed object, with branch-only renders used for inspection. We report PSNR, SSIM, and LPIPS following supplementary Sec.~3.

\begin{figure*}[!t]
\centering
\includegraphics[width=\textwidth]{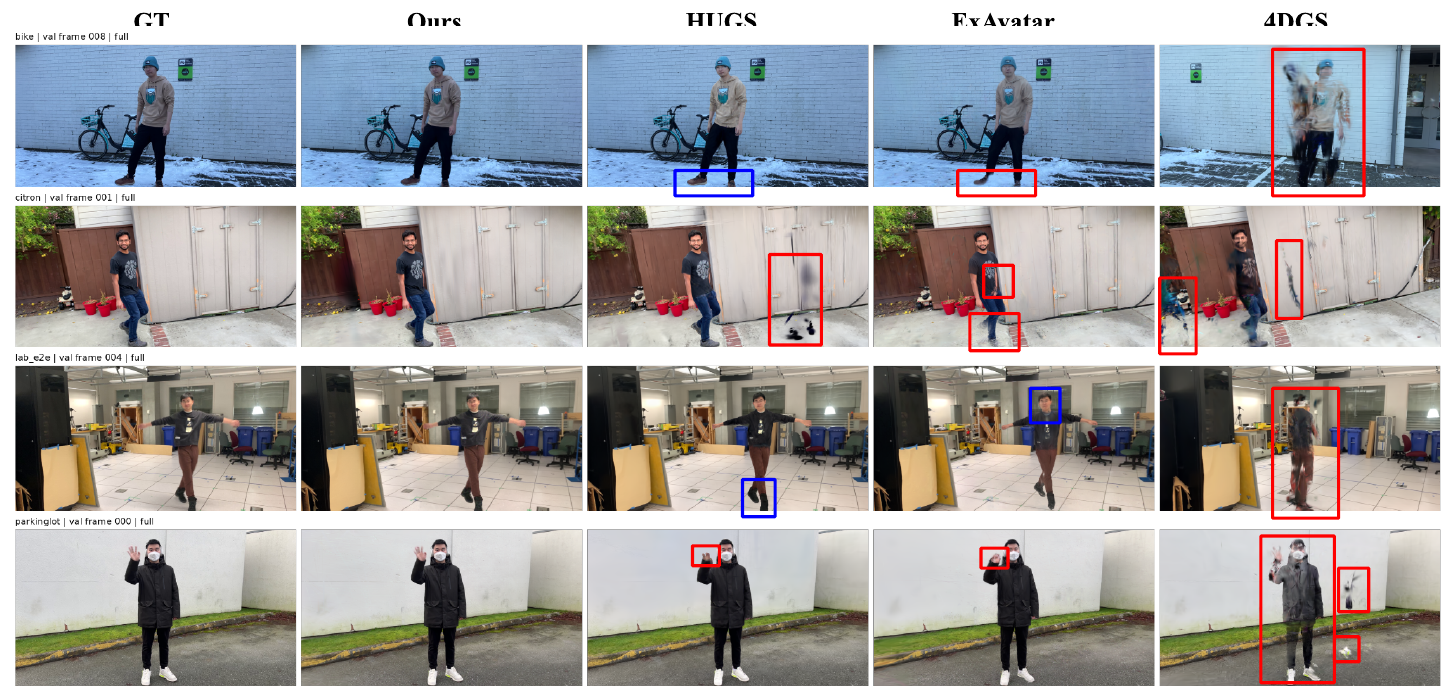}
\captionof{figure}{\textbf{Qualitative comparison on NeuMan validation frames.}
    We compare \proposed with some SOTA models on frames from four NeuMan sequences.
    \textcolor{red}{\textbf{Red}} boxes indicate general reconstruction artifacts such as ghosting, geometry leakage, missing body structure, or inconsistent foreground placement, while \textcolor{blue}{\textbf{blue}} boxes highlight color-intensity errors where the rendering is noticeably lighter or darker than the ground truth.
    Across all examples, \proposed remains closest to the GT in body shape, appearance, and scene consistency.}
\label{fig:simgs_qual}
\end{figure*}

\begin{table}[t]
\centering
\small
\setlength\tabcolsep{4pt}
\caption{\textbf{Prior-initialization ablation.} Ablation on prior initialization over NeuMan dataset. Removing the canonical human seed leaves monocular evidence underconstrained in weakly observed body regions. $\Delta$LPIPS is measured relative to full \proposed.}
\label{tab:ablation_seed}
\resizebox{\columnwidth}{!}{%
\begin{tabular}{l|cccc}
\toprule
\textbf{Variant} & PSNR$\uparrow$ & SSIM$\uparrow$ & LPIPS$\downarrow$ & $\Delta$LPIPS \\
\midrule
Full model (\proposed) & 32.87 & 0.972 & 0.018 & -- \\
w/o human prior seed & 28.73 & 0.918 & 0.044 & +0.026 \\
w/o object seed & 30.45 & 0.931 & 0.047 & +0.029 \\
\bottomrule
\end{tabular}%
}
\vspace{-2mm}
\end{table}

For numerical stability, we use standard implementation guards: scale clamping, early XYZ warmup, seed-rotation repair, and safe handling of empty rasterization cases.

\begin{table}[t]
\centering
\small
\setlength\tabcolsep{3pt}
\caption{\textbf{Branch-specific loss ablation.} Ablation on branch-specific losses and delayed scene-planar regularization. These variants evaluate whether the human foreground losses and scene-planar regularization each contribute to the final compositional reconstruction. $\Delta$LPIPS measures the change from full \proposed.}
\label{tab:ablation_losses}
\resizebox{\columnwidth}{!}{%
\begin{tabular}{l|cccc}
\toprule
\textbf{Variant} & PSNR$\uparrow$ & SSIM$\uparrow$ & LPIPS$\downarrow$ & $\Delta$LPIPS$\downarrow$ \\
\midrule
Full baseline (\proposed) & 29.27 & 0.934 & 0.049 & -- \\
w/o delayed scene-planar regularization & 29.41 & 0.934 & 0.050 & +0.001 \\
w/o human alpha/boundary losses & 25.65 & 0.918 & 0.060 & +0.011 \\
\bottomrule
\end{tabular}%
}
\vspace{-2mm}
\end{table}

\subsection{Main Experiments}

\begin{figure}[t]
\centering
\includegraphics[width=\columnwidth]{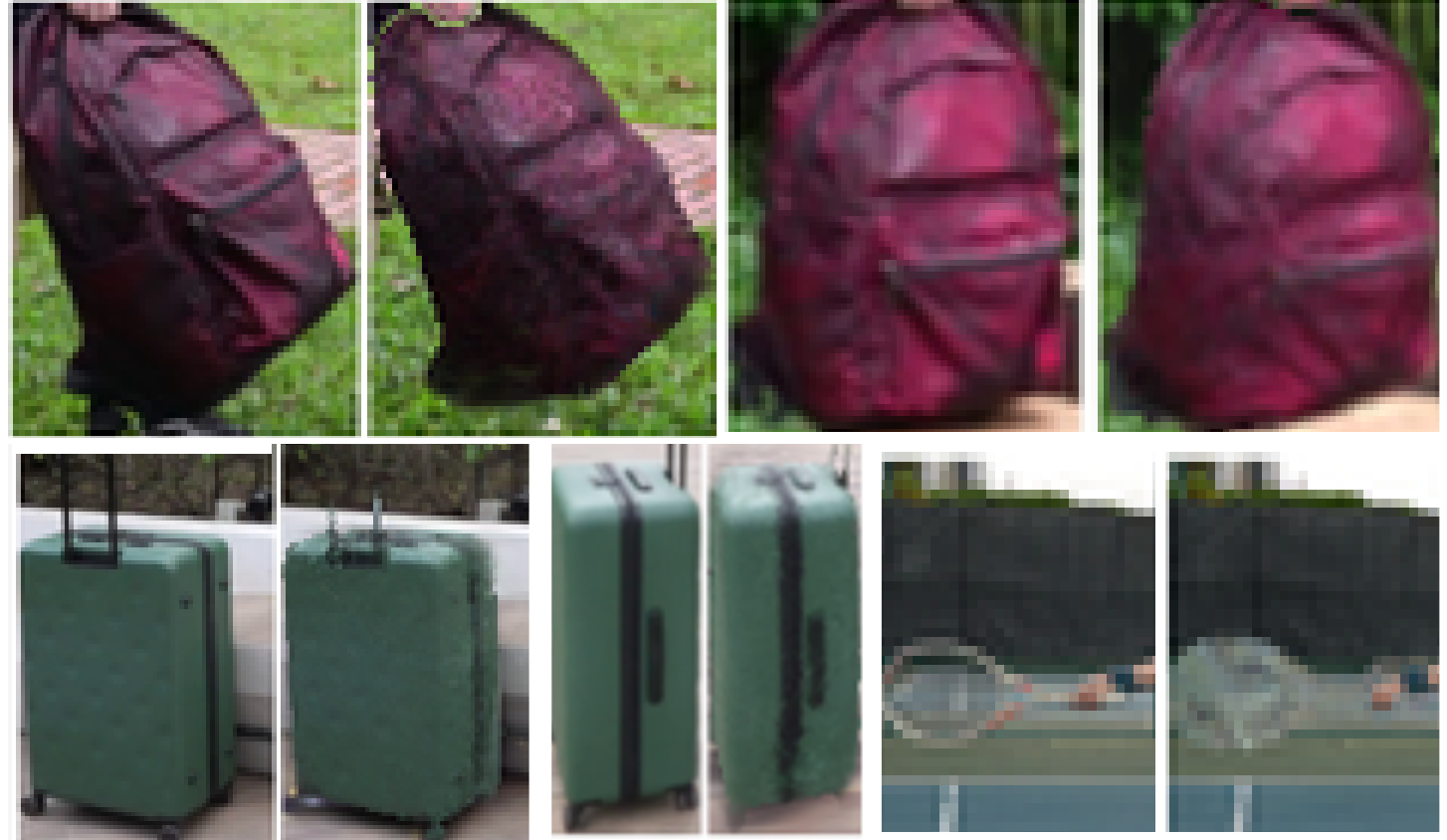}
\caption{\textbf{Object-level qualitative comparison.}
Each pair shows the ground-truth object on the left and the corresponding \proposed rendering on the right.
\proposed preserves object shape, color, and contact-consistent appearance across held-out examples, with the thin tennis racket remaining the hardest case.}
\label{fig:object_qual}
\end{figure}

\paragraph{HOSNeRF.} Table~\ref{tab:hosnerf_comparison} evaluates \proposed on six monocular human--object interaction sequences from HOSNeRF. Among open and reproducible baselines, \proposed obtains the best PSNR on all six scenes and the best LPIPS on five of six scenes. The gains span object occlusion in \textit{Backpack} and \textit{Suitcase}, fast articulated motion in \textit{Tennis} and \textit{Dance}, and larger scene context in \textit{Playground} and \textit{Lounge}. Object crops in Fig.~\ref{fig:object_qual} further show that the object branch preserves manipulated-object appearance in held-out frames. These results support the core design: explicit object motion and delayed object densification sharpen manipulated objects, while visibility-aware human anchoring prevents object appearance from leaking into the body or background.

\paragraph{NeuMan.} Tables~\ref{tab:neuman_human_scene} and~\ref{tab:neuman_human} test whether the human--scene subset of the same compositional representation also improves in-the-wild reconstruction when no manipulated object is present. On full-frame NeuMan rendering, \proposed outperforms NeRF-T, HyperNeRF, Vid2Avatar, NeuMan, and HUGS on every reported scene and metric, improving PSNR by $8.46$--$16.46\%$ and reducing LPIPS by $40.91$--$61.54\%$ over the strongest previous result per sequence. The qualitative comparison in Fig.~\ref{fig:simgs_qual} shows the same trend visually, with \proposed reducing ghosting, body leakage, and color-intensity artifacts. The human-only crop isolates the dynamic person from the easier static background; \proposed again ranks first across all six sequences, with LPIPS reductions of $66.67$--$81.25\%$ on several scenes. This shows that the visibility-aware prior improves human boundaries, limb appearance, and partially observed regions even under moving-camera capture.

These patterns are consistent with the design choices in the method. Full-frame gains come from assigning static background, articulated human motion, and object motion to separate branches, while the larger human-only gains reflect the visibility-aware prior preserving unobserved body regions instead of overfitting them to sparse monocular evidence. The smaller but consistent ablation gaps indicate that no single cue is sufficient alone: initialization, foreground losses, and delayed scene regularization each remove a different failure mode, motivating the limitations and future directions summarized in the conclusion.

\subsection{Ablation Studies}
\label{sec:ablations}

% We report the highest-impact initialization and loss ablations in the main paper; additional object-seed, visibility-anchor, and schedule ablations are provided in supplementary material. Each ablation removes a source of branch reliability and exposes the ambiguity it resolves during monocular optimization.

\paragraph{Impact of human prior initialization.}
Table~\ref{tab:ablation_seed} shows that removing either human or object initialization hurts perceptual quality, increasing LPIPS by $+0.026$ and $+0.029$, respectively. The human prior is most useful in self-occluded or object-occluded regions where RGB and mask supervision are weak; without it, visible pixels can still be fitted, but hidden structure and boundaries become less coherent. The object seed similarly reduces scale, depth, and shape ambiguity from masks alone, making later object-motion refinement less prone to drifting into the human or scene branch.

\paragraph{Impact of branch-specific losses.}
Table~\ref{tab:ablation_losses} shows that human alpha/boundary losses are important for foreground quality, while delayed scene-planar regularization mainly improves structural consistency. Removing human foreground losses causes the larger drop because the compositor loses an explicit signal separating limbs from nearby scene and object pixels, increasing branch leakage. Removing delayed scene-planar regularization has a smaller numerical effect because repeated views already supervise the background, but planar cues still discourage fragmentation of large static surfaces after photometric fitting becomes reliable.

%% file: sections/conclusion.tex
\section{Conclusion and Future Work}

We presented \proposed, a compositional Gaussian-splatting framework for reconstructing dynamic human--object scenes from monocular video. By separating the scene into articulated human, rigid object, and static background branches, then fusing them with staged optimization, visibility-aware anchoring, object-specific supervision, and delayed scene-planar regularization, \proposed improves full-scene and human-focused reconstruction on HOSNeRF and NeuMan under occlusion, object contact, and limited camera coverage. Its main limitations are dependence on monocular preprocessing, imperfect recovery of fully unobserved identity or clothing detail, weak rather than complete scene geometry from planar primitives, and the rigid-object assumption. Future work should jointly refine preprocessing and reconstruction, support deformable or articulated objects, and strengthen scene geometry.